\documentclass[10pt,twocolumn,letterpaper]{article}

\usepackage{cvpr}
\usepackage{times}
\usepackage{epsfig}
\usepackage{graphicx}
\usepackage{amsmath}
\usepackage{amssymb}

\usepackage[tableposition=above]{caption}
\usepackage{bbm}
\usepackage{bm}
\usepackage{xspace}
\usepackage[utf8]{inputenc} 
\usepackage{url}            
\usepackage{booktabs}       
\usepackage{amsfonts}       
\usepackage{nicefrac}       
\usepackage{microtype}      
\usepackage{color}
\usepackage[labelformat=empty]{subfig}
\usepackage{algorithm}
\usepackage{algorithmic}
\usepackage{array}
\usepackage{multirow}
\usepackage[para,online,flushleft]{threeparttable}
\usepackage{float}
\usepackage{enumitem}
\usepackage{makecell}
\usepackage{gensymb}         
\usepackage{mathtools}
\usepackage{capt-of}
\usepackage{amssymb}
\usepackage{caption}
\usepackage[normalem]{ulem}

\usepackage{xcolor}
\usepackage{ifthen}
\usepackage{textcomp}
\definecolor{red}{rgb}{0.9,0.1,0}
\definecolor{slateblue}{rgb}{0.7,0.35,0.9}
\definecolor{green}{rgb}{0, 0.4, 0}
\definecolor{brown}{rgb}{0.3, 0.2, 0}
\definecolor{mahogany}{rgb}{0.75, 0.25, 0.0}
\definecolor{purple}{rgb}{0.3, 0, 0.3}
\definecolor{darkgreen}{rgb}{0, 0.4, 0}
\definecolor{frenchblue}{rgb}{0.0, 0.45, 0.73}
\definecolor{blue}{rgb}{0.0, 0.0, 1.0}
\definecolor{goldenrod}{rgb}{0.65, 0.45, 0.03}
\definecolor{gray}{rgb}{0.5,0.5,0.5}

\newboolean{revising}
\setboolean{revising}{true}
\ifthenelse{\boolean{revising}}
{
    
    \newcommand{\first}[1]{\textcolor{red}{#1}}
    \newcommand{\second}[1]{\textcolor{blue}{#1}}

} {

}

\newcommand\blfootnote[1]{%
  \begingroup
  \renewcommand\thefootnote{}\footnote{#1}%
  \addtocounter{footnote}{-1}%
  \endgroup
}

\usepackage[breaklinks=true,bookmarks=false]{hyperref}

\cvprfinalcopy 

\def\cvprPaperID{2915} 

\begin{document}

\title{CLCC: Contrastive Learning for Color Constancy}


\author{
Yi-Chen Lo*,
Chia-Che Chang*,
Hsuan-Chao Chiu,
Yu-Hao Huang, \\
Chia-Ping Chen,
Yu-Lin Chang,
Kevin Jou \\

MediaTek Inc., Hsinchu, Taiwan \\
{\tt\small \{yichen.lo, chia-che.chang, Hsuanchao.Chiu, justin-yh.huang,} \\ 
{\tt\small chiaping.chen, yulin.chang, kevin.jou\}@mediatek.com}

}

\maketitle

\blfootnote{* Indicates equal contribution.}

\thispagestyle{empty}
\pagestyle{empty}

\begin{abstract}
In this paper, we present CLCC, a novel contrastive learning framework for color constancy. Contrastive learning has been applied for learning high-quality visual representations for image classification.
One key aspect to yield useful representations for image classification is to design illuminant invariant augmentations.
However, the illuminant invariant assumption conflicts with the nature of the color constancy task, which aims to estimate the illuminant given a raw image.
Therefore, we construct effective contrastive pairs for learning better illuminant-dependent features via a novel raw-domain color augmentation.
On the NUS-8 dataset, our method provides $17.5\%$ relative improvements over a strong baseline, reaching state-of-the-art performance without increasing model complexity. Furthermore, our method achieves competitive performance on the Gehler dataset with $3\times$ fewer parameters compared to top-ranking deep learning methods.
More importantly, we show that our model is more robust to different scenes under close proximity of illuminants, significantly reducing $28.7\%$ worst-case error in data-sparse regions. Our code is available at \url{https://github.com/howardyclo/clcc-cvpr21}.
\end{abstract}


\begin{figure}[t]
\begin{center}
   \includegraphics[width=0.8\linewidth]{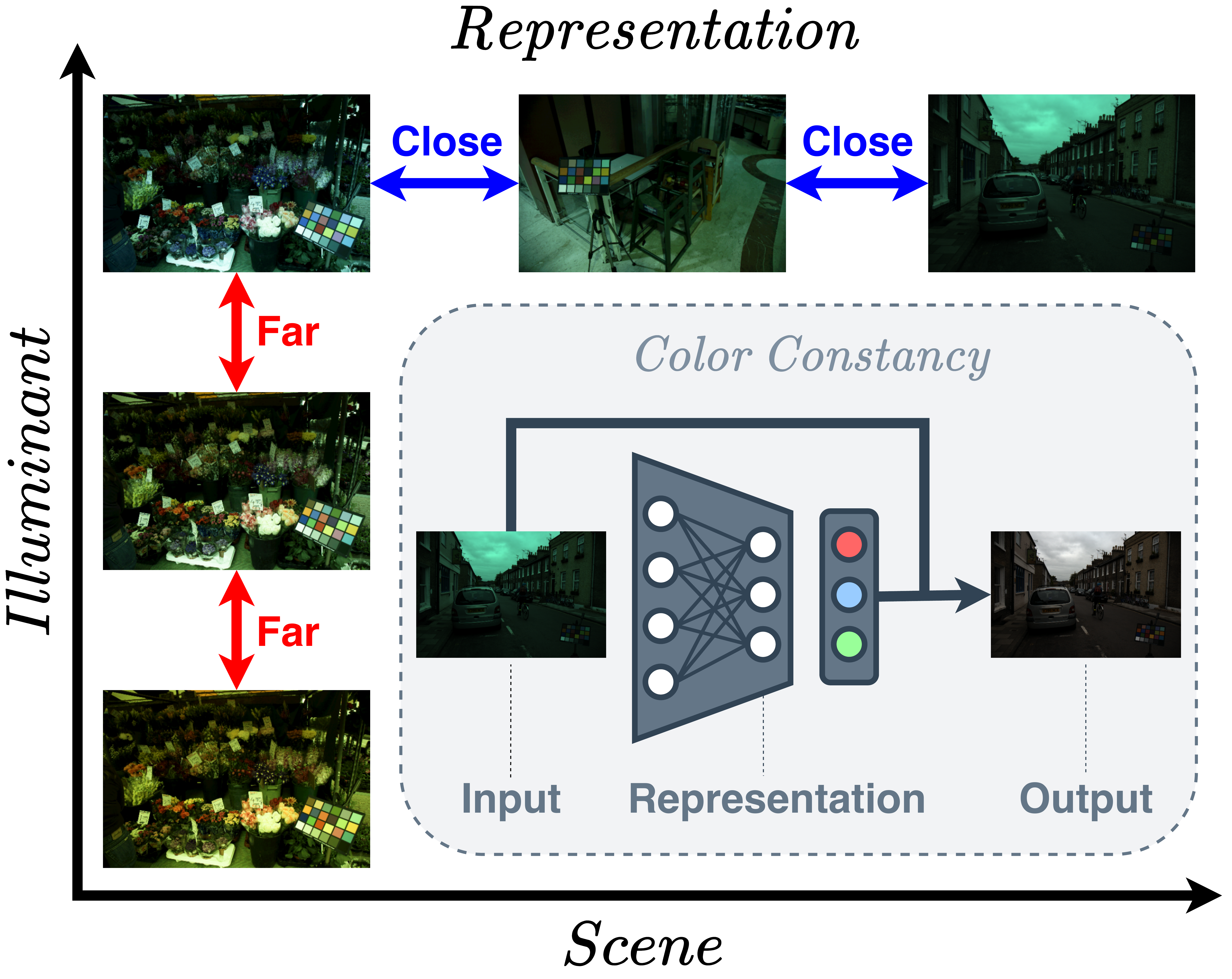}
\end{center}
\vspace*{-5mm}
\caption{Our main idea of CLCC: The scene-invariant, illuminant-dependent representation of the same scene under different illuminants should be far from each other, while different scenes under the same illuminant should be close to each other.}
\label{fig:motivation_v2}
\end{figure}

\section{Introduction}
The human visual system can perceive the same canonical color of an object even under different illuminants. 
This feature can be mimicked by computational color constancy, an essential task in the camera pipeline that processes raw sensor signals to sRGB images.
Conventional methods~\cite{buchsbaum1980spatial,FinlaysonT04,FuntS10,land1971lightness,WeijerGG07} utilize statistical properties of the scene to cope with this ill-posed problem, such as the most widely used gray world assumption.
Such statistical methods, however, often fail where their assumptions are violated in complex scenes. 

Until recently, deep learning based methods~\cite{HuWL17,Qian0NKM17,XuLHLQ20,YuCWQZJ20} have been applied to the color constancy problem and achieve considerable quality improvements on challenging scenes.
Yet, this ill-posed and sensor-dependent task still suffers from the difficulty of collecting massive paired data for supervised training.

When learning with insufficient training data, a common issue frequently encountered is the possibility of learning spurious correlations~\cite{Vigen} or undesirable biases from data~\cite{TorralbaE11}: misleading features that work for most training samples but do not always hold in general.
For instance, previous research has shown that a deep object-recognition model may rely on the spuriously correlated background instead of the foreground object to make predictions~\cite{abs-2006-09994} or be biased towards object textures instead of shapes~\cite{GeirhosRMBWB19}.
In the case of color constancy, outdoor scenes often have higher correlations with high color temperature illuminants than indoor scenes.
Thus, deep learning models may focus on scene related features instead of illuminant related features. This leads to a decision behavior that tends to predict high color temperature illuminants for outdoor scenes, but suffers high error on outdoor scenes under low color temperature illuminants.
This problem becomes worse when the sparsity of data increases.

To avoid learning such spurious correlations, one may seek to regularize deep learning models to learn scene-invariant, illuminant-dependent representations. 
As illustrated in Fig.\ref{fig:motivation_v2}, in contrast to image classification problem, the representation of the same scene under different illuminants should be \emph{far} from each other. 
On the contrary, the representation of different scenes under the same illuminant should be \emph{close} to each other. 
Therefore, we propose to learn such desired representations by contrastive learning~\cite{simclr,He0WXG20,HjelmFLGBTB19}, a framework that learns general and robust representations by comparing similar and dissimilar samples.

However, conventional self-supervised contrastive learning often generates easy or trivial contrastive pairs that are not very useful for learning generalized feature representations~\cite{SupContrastKhosla}. 
To address this issue, a recent work~\cite{simclr} has demonstrated that strong data augmentation is crucial for conducting successful contrastive learning.

Nevertheless, previous data augmentations that have been shown effective for image classification may not be suitable for color constancy.
Here we illustrate some of them. 
First, most previous data augmentations in contrastive learning are designed for high-level vision tasks (e.g., object recognition) and seek illuminant invariant features, which can be detrimental for color constancy.
For example, color dropping converts an sRGB image to a gray-scale one, making the color constancy task even more difficult.
Moreover, the color constancy task works best in the linear color space where the linear relationship to scene radiance is preserved. 
This prevents from using non-linear color jittering augmentations, e.g., contrast, saturation, and hue.

To this end, we propose \textit{CLCC: Contrastive Learning for Color Constancy}, a novel color constancy framework with contrastive learning. 
For the purpose of color constancy, effective positive and negative pairs are constructed by exploiting the label information, 
while novel color augmentations are designed based on color domain knowledge~\cite{AfifiB19iccv,RawtorawBrown14,ColorReprKaraimerB18}.

Built upon a previous state-of-the-art~\cite{HuWL17}, CLCC provides additional 17.5$\%$ improvements (mean angular error decreases from 2.23 to 1.84) on a public benchmark dataset~\cite{cheng2014illuminant}, achieving state-of-the-art results without increasing model complexity. 
Besides accuracy improvement, our method also allows deep learning models to effectively acquire robust and generalized representations even when learning from small training datasets.


\paragraph{Contribution}
We introduce CLCC, a fully supervised contrastive learning framework for the task of color constancy.
By leveraging label information, CLCC generates more diverse and harder contrastive pairs to effectively learn feature representations aiming for better quality and robustness. 
A novel color augmentation method that incorporates color domain knowledge is proposed. 
We improve the previous state-of-the-art deep color constancy model without increasing model complexity. 
CLCC encourages learning illuminant-dependent features rather than spurious scene content features irrelevant for color constancy, making our model more robust and generalized, especially in data-sparse regions. 
\section{Related Work}
\subsection{Contrastive learning}
Contrastive learning is a framework that learns general and robust feature representations by comparing similar and dissimilar pairs.
Inspired from noise contrastive estimation (NCE) and N-pair loss~\cite{GutmannH10,MnihK13,Sohn16}, remarkable improvements on image classification are obtained in several recent works~\cite{simclr,He0WXG20,HjelmFLGBTB19,LoweOV19,abs-1906-05849,abs-1807-03748,WuXYL18}.
Particularly, a mutual information based contrastive loss, InfoNCE~\cite{abs-1807-03748} has become a popular choice for contrastive learning (see \cite{McAllesterS20,PooleOOAT19} for more discussion).
Furthermore, recent works~\cite{BachmanHB19,BeyerZOK19,ChuangRL0J20,abs-1905-09272,SupContrastKhosla,abs-2005-10243} have shown that leveraging supervised labels not only improves learning efficiency by alleviating sampling bias (and hence reducing the need for large batch size training) but also improves generalization by learning task-relevant features.

\subsection{Data augmentation}
Data augmentations such as random cropping, flipping, and rotation have been widely used in classification ~\cite{HeZRS16,SandlerHZZC18}, object detection~\cite{LinGGHD17}, and semantic segmentation~\cite{ChenZPSA18} to improve model quality.
Various works rely on manually designed augmentations to reach their best results~\cite{simclr,SatoNY15}. To ease such efforts, strategy search~\cite{CubukZMVL19,CubukZSL20} or data synthesis~\cite{abs-1712-04621,abs-1711-00648} have been used to improve data quality and diversity. 
However, popular data augmentation strategies for image recognition~\cite{simclr,CubukZMVL19,KalantariR17,RedmonDGF16} (e.g., color channel dropping, color channel swapping, HSV jittering) may not be suitable for the color constancy task.
Thus, we incorporate color domain knowledge~\cite{AfifiB19iccv,ColorReprKaraimerB18,RawtorawBrown14} to design data augmentation suitable for contrastive learning on color constancy.

\subsection{Color constancy}

Color constancy is a fundamental low-level computer vision task that has been studied for decades. In general, current research can be divided into learning-free and learning-based approaches.
The former ones use color histogram and spatial information to estimate illuminant~\cite{buchsbaum1980spatial,FinlaysonT04,FuntS10,land1971lightness,WeijerGG07}. 
Despite the efficiency of these methods, they do not perform well on  challenging scenes with ambiguous color pixels.
The latter ones adopt data-driven approaches that learn to estimate illuminant from training data~\cite{Barnard00,BarronT17,Finlayson13,FuntX04,JozeD12}. These learning-based approaches outperform learning-free methods and have become popular in both academic and industry fields. In addition, recent works have shown that features learned from deep neural networks are better than hand-crafted ones~\cite{KrizhevskySH12,LiYWZH18,RedmonDGF16}.
Consequently, deep learning based color constancy research has gradually received more and more attention. Recently, FC4 uses ImageNet-pretrained backbones~\cite{HuWL17,KrizhevskySH12} to prevent over-fitting and estimate illuminant with two additional convolutional layers. RCC-Net~\cite{Qian0NKM17} uses a convolutional LSTM to extract features in both spatial and temporal domains to estimate illuminants.
C4~\cite{YuCWQZJ20} proposes a cascaded, coarse-to-fine network for color constancy, stacking three SqueezeNets to improve model quality.
To mitigate the issue that the learned representation suffers from being sensitive to image content, IGTN ~\cite{XuLHLQ20} introduces metric learning to learn scene-independent illuminant features.
From a different perspective, most learning based methods strongly bind to a single sensor's spectral sensitivity and thus cannot be generalized to other camera sensors without fine-tuning.
Several works~\cite{AfifiB19,JuarezPBLSM20,XiaoG020} have attempted to resolve this issue by training on multiple sensors simultaneously. We note that multi-sensor training is out of the scope of this work, hence we do not compare to this line of research.

%
\begin{figure}[t]
\begin{center}
  \includegraphics[width=1.0\linewidth]{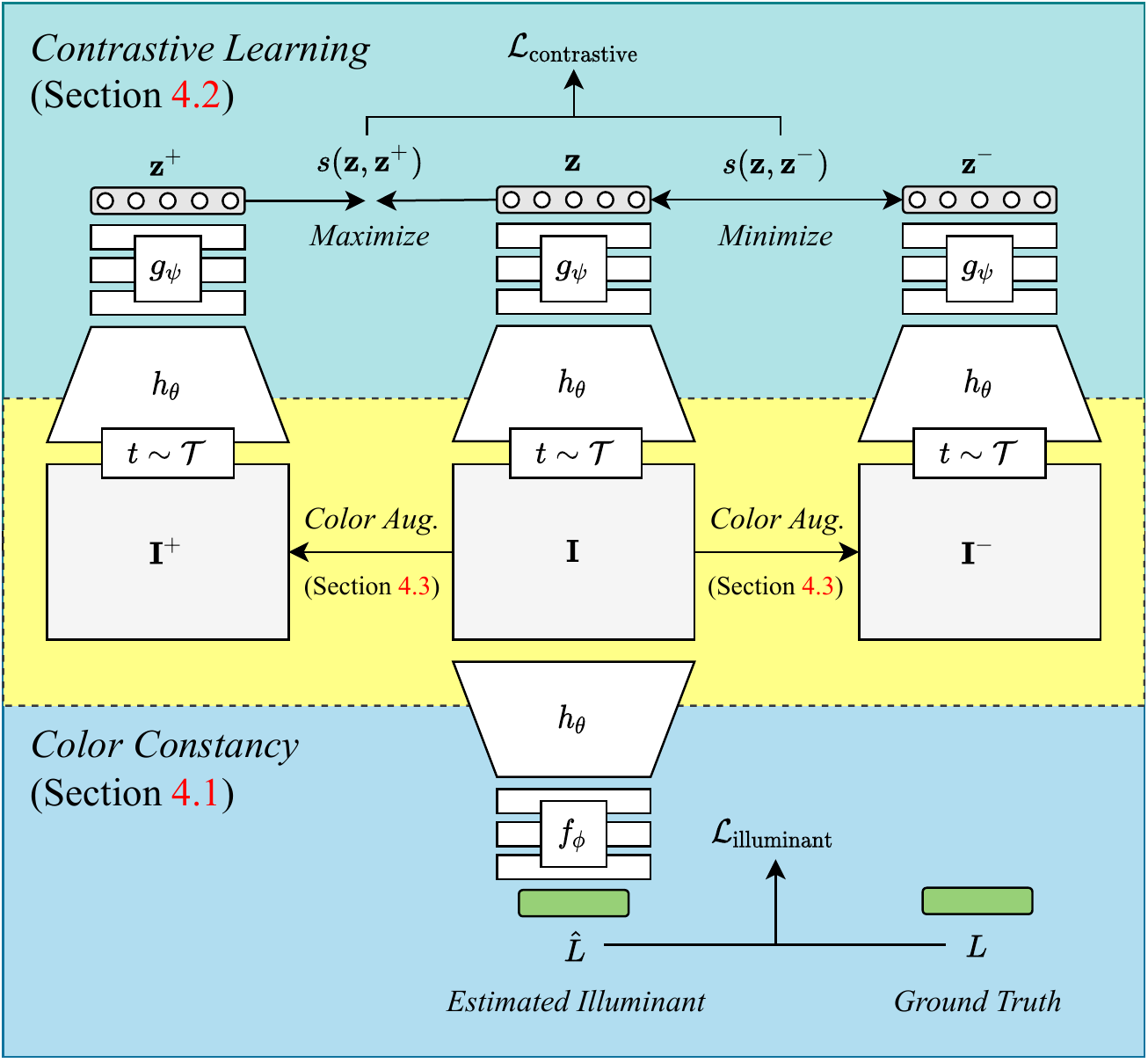}
\end{center}
\vspace*{-5mm}
  \caption{An overview of our CLCC: Besides the main color constancy task, we propose to incorporate contrastive learning to learn generalized and illuminant-dependent feature representations.}
\label{fig:overview_architecture}
\end{figure}
\section{Preliminaries}
\paragraph{Image formation model}
A raw-RGB image can be viewed as a measurement of scene radiance within a particular range of spectrum from a camera sensor:
\begin{align}
\mathbf{I}_{\mathrm{raw}}(\mathbf{x})=\int_{\omega} R_{c}(\lambda) S(\mathbf{x}, \lambda) L(\lambda) d \lambda
\label{eqn:imgformationmodel}
\end{align}
where $\lambda$ denotes the wavelength, $\omega \in [380, 720]$ (nm) is the visible spectrum, $R_{c}$ is the spectral sensitivities of the sensor's color channel $c \in \{r,g,b\}$. The term $S(\mathbf{x}, \lambda)$ denotes the scene's material reflectance at pixel $\mathbf{x}$ and $L(\lambda)$ is the illuminant in the scene, assumed to be spatially uniform. Notably, $\mathbf{I}_{\mathrm{raw}}$ values are linearly proportional to the scene radiance, making color constancy easier to work with.
\paragraph{Color space conversions}
Usually $\mathbf{I}_{\mathrm{raw}}$ undergoes two color space conversions in the camera pipeline:
\begin{align}
\mathbf{I}_{\mathrm{sRGB}} = \mathcal{G}_{\mathrm{XYZ} \rightarrow \mathrm{sRGB}} (
    \mathcal{F}_{\mathrm{raw} \rightarrow \mathrm{XYZ}}(
        \mathbf{I}_{\mathrm{raw}}
    )
)
\label{eqn:colorspaceconversion}
\end{align}
where $\mathcal{F}(\cdot)$ involves linear operations including white balance and full color correction. $\mathcal{F}(\cdot)$ maps a sensor-specific raw-RGB to a standard perceptual color space such as CIE XYZ. $\mathcal{G}(\cdot)$ involves non-linear photo-finishing procedures (e.g., contrast, hue, saturation) and eventually maps XYZ to the sRGB color space (we refer to \cite{Karaimer2016ASP} for a complete overview of camera imaging pipeline).
\paragraph{White balance and full color correction}
Given $\mathbf{I}_{\mathrm{raw}}$, white balance (WB) aims to estimate the scene illuminant $L = [L_r, L_g, L_b]$, i.e., the color of a neutral material captured with a physical color checker placed in the scene. Knowing that a neutral material equally reflects spectral energy at every wavelength regardless of different illuminants, we can apply a $3\times3$ diagonal matrix $\mathbf{M}_{\mathrm{WB}}$ with the diagonal entries $[L_g/L_r, 1, L_g/L_b]$ on $\mathbf{I}_{\mathrm{raw}}$ to obtain a white-balanced image $\mathbf{I}_{\mathrm{WB}}$:
\begin{align}
\mathbf{I}_{\mathrm{WB}} = \mathbf{I}_{\mathrm{raw}} \mathbf{M}_{\mathrm{WB}}
\label{eqn:WB}
\end{align}
After WB, a neutral material should appear achormatic (i.e., ``gray'').
Because WB only corrects achromatic colors, a $3\times3$ full color correction matrix $\mathbf{M}_{\mathrm{CC}}$ is further applied to correct chromatic colors (in practice, those chromatic patches with known CIE XYZ values on color checker). Note that $\mathbf{M}_{\mathrm{CC}}$ is illuminant-specific due to error introduced by the estimated $\mathbf{M}_{\mathrm{WB}}$
\begin{align}
\mathbf{I}_{\mathrm{XYZ}} = \mathbf{I}_{\mathrm{WB}} \mathbf{M}_{\mathrm{CC}}
\label{eqn:CC}
\end{align}
Such $\mathbf{I}_{\mathrm{XYZ}}$ is sensor-agnostic since the illuminant cast is completely removed for both achromatic and chromatic colors.
\section{Methodology}
We start with our problem formulation and review conventional self-supervised contrastive learning in Section \ref{sec:formulation}. Next, we introduce CLCC,  our fully-supervised contrastive learning framework for color constancy in Section \ref{sec:clcc}. Finally, we describe our color augmentation for contrastive pair synthesis in Section \ref{sec:coloraug}. How these sections fit together is illustrated in Fig.~\ref{fig:overview_architecture}.
\begin{figure*}
\begin{center}
   \includegraphics[width=1.0\linewidth]{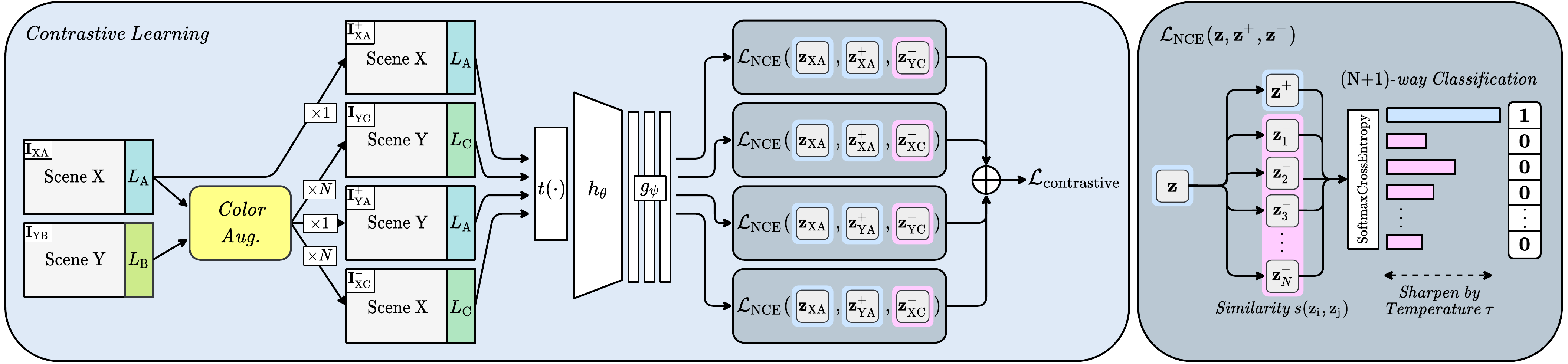}
\end{center}
\vspace*{-5mm}
   \caption{The proposed formation for contrastive pairs and color augmentation.}
\label{fig:clcc}
\end{figure*}
\subsection{Formulation}
\label{sec:formulation}
\paragraph{The learning problem}
Our problem setting follows the majority of learning-based color constancy research which only focuses on the white balance step of estimating the illuminant $L$ from the input raw image $\mathbf{I}_{\mathrm{raw}}$:
\begin{align}
\hat{L} = f_{\phi}(h_{\theta}(
\mathbf{I}_{\mathrm{raw}}
))
\label{eqn:problemformulation}
\end{align}
where $h_{\theta}$ is the feature extractor that produces visual representations for $\mathbf{I}_{\mathrm{raw}}$, $f_{\phi}$ is the illuminant estimation function, and $\hat{L}$ is the estimated illuminant. Both $h_{\theta}$ and $f_{\phi}$ are parameterized by deep neural networks with arbitrary architecture design, where $\theta$ and $\phi$ can be trained via back-propagation.
\paragraph{The learning objectives}
The overall learning objective can be decomposed into two parts: (1) illuminant estimation for color constancy and (2) contrastive learning for better representations (as shown in Fig.~\ref{fig:overview_architecture}): 
\begin{align}
\mathcal{L}_{\mathrm{total}} =
\lambda \mathcal{L}_{\mathrm{illuminant}} +
\beta \mathcal{L}_{\mathrm{contrastive}}
\label{eqn:learningobjective}
\end{align}
For the illuminant estimation task, we use the commonly used angular error as:
\begin{align}
\mathcal{L}_{\mathrm{illuminant}} =
\arccos{(\frac
    {\hat{L} \cdot L}
    {\lVert \hat{L} \rVert \cdot \lVert L \rVert}
)}
\label{eqn:angularloss}
\end{align}
where $\hat{L}$ is the estimated illuminant and $L$ is the ground-truth illuminant.

Since the datasets for color constancy are  relatively small because it is difficult to collect training data with corresponding ground-truth illuminants.
Training a deep learning model with only the supervision $\mathcal{L}_{\mathrm{illuminant}}$ usually does not generalize well.
Therefore, we propose to use contrastive learning, which can help to learn a color constancy model that generalize better even with a small training dataset.  
Details of the contrastive learning task are described as follows.
\paragraph{The contrastive learning framework}
The proposed CLCC is built upon the recent work SimCLR~\cite{simclr}.
Therefore, we discuss self-supervised constrative learning for color constancy in this section, and then elaborate on our extended fully-supervised contrastive learning in the next section. 
The essential building blocks of contrastive learning are illustrated here:
\begin{itemize}
    \setlength \itemsep{-0.0em}
    \item A stochastic data augmentation $t(\cdot) \sim \mathcal{T}$ that augments a sample image $\mathbf{I}$ to a different \textit{view} $t(\mathbf{I})$. Note that $t(\cdot)$ is required to be \textit{label-preserving}, meaning that $\mathbf{I}$ and $t(\mathbf{I})$ still share the same ground-truth illuminant $L$.
    \item A feature extraction function $h_\theta$ that extracts the \textit{representation} of $t(\mathbf{I})$. $h_\theta$ is further used for downstream color constancy task as defined in the Eq.~(\ref{eqn:problemformulation}).
    \item A feature projection function $g_\psi$ that maps the representation  $h_{\theta}(t(\mathbf{I}))$ to the \textit{projection} $\mathbf{z}$ that lies on a unit hypersphere.
    $g_\psi$ is typically only required when learning representations and thrown away once the learning is finished.
    \item A similarity metric function $s(\cdot)$ that measures the similarity between latent projections $(\mathbf{z}_i, \mathbf{z}_j)$.
    \item Contrastive pair formulation: \textit{anchor} $\mathbf{I}$, \textit{positive} $\mathbf{I}^{+}$ and \textit{negative} $\mathbf{I}^{-}$ samples jointly compose the positive pair $(\mathbf{I}, \mathbf{I}^{+})$ and the negative pair $(\mathbf{I}, \mathbf{I}^{-})$ for contrastive learning. For the color constancy task, a positive pair should share the same illuminant label $L$, while a negative pair should have different ones.
    \item A contrastive loss function $\mathcal{L}_{\mathrm{contrastive}}$ that aims to maximize the similarity between the projection of the positive pair $(\mathbf{z}, \mathbf{z}^{+})$ and minimize the similarity between that of the negative pair $(\mathbf{z}, \mathbf{z}^{-})$ in the latent projection space.
\end{itemize}
\paragraph{Self-supervised contrastive learning}
Given two random training images $\mathbf{I}_i$ and $\mathbf{I}_j$ with different scene content, one can naively form a positive contrastive pair with two randomly augmented views of the same image $(t(\mathbf{I}_i), t'(\mathbf{I}_i^{+}))$, and a negative contrastive pair with views of two different images $(t(\mathbf{I}_i), t'(\mathbf{I}_j^{-}))$. 

Such naive formulation introduces two potential drawbacks.
One is the \textit{sampling bias}, the potential to sample a false negative pair that shares very similar illuminants (i.e., $L_i \simeq L_j$).
The other is the \textit{lack of hardness}, the fact that the positive  $t(\mathbf{I}_i^{+})$ derived from the same image as the anchor $t(\mathbf{I}_i)$ could share similar scene content. This alone suffices to let neural networks easily distinguish from negative $t'(\mathbf{I}_j^{-})$ with apparently different scene content. Hence, as suggested by~\cite{simclr}, one should seek strong data augmentations to regularize such learning shortcut.

To alleviate sampling bias and increase the hardness of contrastive pairs, we propose to leverage label information, extending self-supervised contrastive learning into fully-supervised contrastive learning, where the essential data augmentation is specifically designed to be label-preserving for color constancy task.

\subsection{\textbf{\textit{CLCC}}: Contrastive learning for color constancy}
\label{sec:clcc}
We now describe our realization of each component in the proposed fully-supervised contrastive learning framework, as depicted in Fig.~\ref{fig:clcc}
\paragraph{Contrastive pair formulation}
Here, we define $\mathbf{I}_{\mathrm{X}\mathrm{A}}$ as a linear raw-RGB image captured in the scene $\mathrm{X}$ under the illuminant $L_{\mathrm{A}}$. Let us recapitulate our definition that a positive pair should share an identical illuminant while a negative pair should not. Therefore, given two randomly sampled training images $\mathbf{I}_{\mathrm{AX}}$ and $\mathbf{I}_{\mathrm{BY}}$,
we construct our contrastive pairs as follows:
\begin{itemize}
    \setlength \itemsep{-0.0em}
    \item An easy positive pair $(t(\mathbf{I}_{\mathrm{XA}}), t'(\mathbf{I}_{\mathrm{XA}}^{+}))$---with an identical scene $\mathrm{X}$ and illuminant $L_{\mathrm{A}}$. 
    \item An easy negative pair $(t(\mathbf{I}_{\mathrm{XA}}), t'(\mathbf{I}_{\mathrm{YC}}^{-}))$---with different scenes ($\mathrm{X}$, $\mathrm{Y}$) and different illuminants ($L_{\mathrm{A}}$, $L_{\mathrm{C}}$).
    \item A hard positive pair $(t(\mathbf{I}_{\mathrm{XA}}), t'(\mathbf{I}_{\mathrm{YA}}^{+}))$---with different scenes ($\mathrm{X}$, $\mathrm{Y}$) but an identical illuminant $L_{\mathrm{A}}$.
    \item A hard negative pair $(t(\mathbf{I}_{\mathrm{XA}}), t'(\mathbf{I}_{\mathrm{XC}}^{-}))$---with an identical scene $\mathrm{X}$ but different illuminants ($L_{\mathrm{A}}$, $L_{\mathrm{C}}$).
\end{itemize}
$\mathbf{I}_{\mathrm{YC}}$, $\mathbf{I}_{\mathrm{YA}}$ and $\mathbf{I}_{\mathrm{XC}}$ are synthesized by replacing one scene's illuminant to another.
Note that we define the novel illuminant $L_{\mathrm{C}}$ as the interpolation or extrapolation between $L_{\mathrm{A}}$ and $L_{\mathrm{B}}$, thus we do not need a redundant hard negative sample $\mathbf{I}_{\mathrm{XB}}$.
More details are explained in Section \ref{sec:coloraug}.
$t$ is a stochastic perturbation-based, illuminant-preserving data augmentation composed by \textit{random intensity}, \textit{random shot noise}, and \textit{random Gaussian noise}.
\paragraph{Similarity metric and contrastive loss function}
Once the contrastive pairs are defined in the image space, we use $h_\theta$ and $g_\psi$ to encode those views $t(\cdot)$ to the latent projection space $\mathbf{z}$. Our contrastive loss can be computed as the sum of InfoNCE losses for properly elaborated contrastive pairs:
\begin{equation}
\begin{aligned}
\mathcal{L}_{\mathrm{contrastive}} &= 
\mathcal{L}_{\mathrm{NCE}}(\mathbf{z}_{\mathrm{XA}},\mathbf{z}_{\mathrm{XA}}^{+},\mathbf{z}_{\mathrm{YC}}^{-})\\ &+
\mathcal{L}_{\mathrm{NCE}}(\mathbf{z}_{\mathrm{XA}},\mathbf{z}_{\mathrm{XA}}^{+},\mathbf{z}_{\mathrm{XC}}^{-})\\ &+
\mathcal{L}_{\mathrm{NCE}}(\mathbf{z}_{\mathrm{XA}},\mathbf{z}_{\mathrm{YA}}^{+},\mathbf{z}_{\mathrm{YC}}^{-})\\ &+ 
\mathcal{L}_{\mathrm{NCE}}(\mathbf{z}_{\mathrm{XA}},\mathbf{z}_{\mathrm{YA}}^{+},\mathbf{z}_{\mathrm{XC}}^{-})
\label{eqn:contrastiveloss}
\end{aligned}
\end{equation}
The InfoNCE loss $\mathcal{L}_{\mathrm{NCE}}$ can be computed as:
\begin{equation}
\begin{aligned}
\mathcal{L}_{\mathrm{NCE}} =
-\log \left[\frac
    {\exp (s^{+} / \tau)}
    {\exp (s^{+} / \tau) + \sum_{n=1}^{N} \exp(s^{-} / \tau)}
\right]
\label{eqn:infonceloss}
\end{aligned}
\end{equation}
where $s^{+}$ and $s^{-}$ are the cosine similarity scores of positive and negative pairs respectively:
\begin{equation}
\begin{aligned}
s^{+} &= s(\mathbf{z}, \mathbf{z}^{+})\\
s^{-} &= s(\mathbf{z}, \mathbf{z}^{-})
\label{eqn:similarity}
\end{aligned}
\end{equation}
Equation \eqref{eqn:infonceloss} could be viewed as performing a $(N+1)$-way classification realized by cross-entropy loss with $N$ negative pairs and $1$ positive pair.
$\tau$ is the temperature scaling factor.
\begin{figure*}
\begin{center}
   \includegraphics[width=1.0\linewidth]{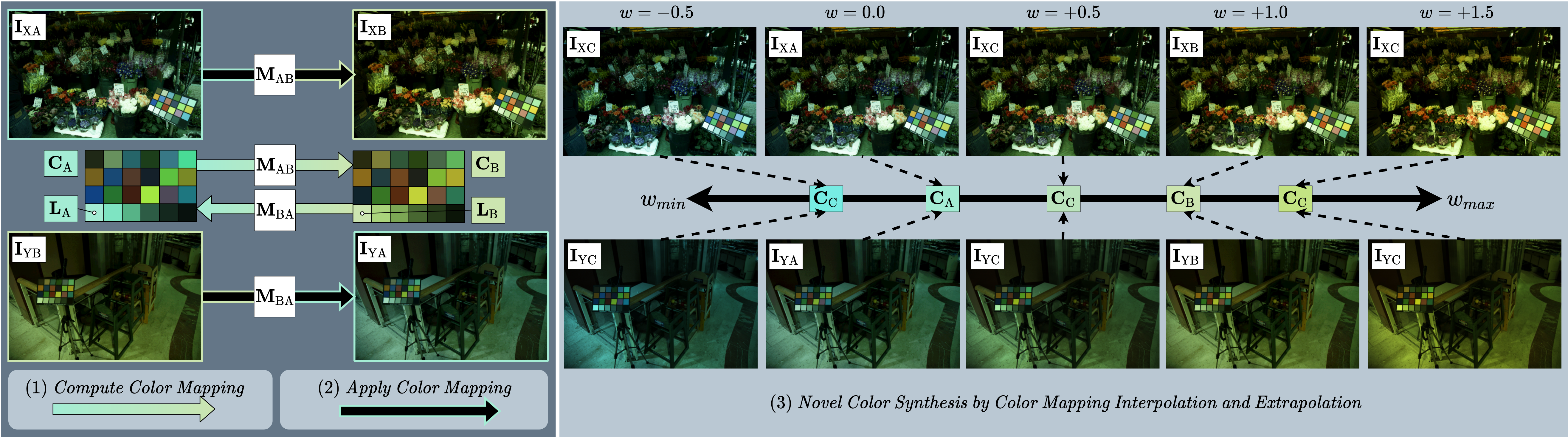}
\end{center}
\vspace*{-5mm}
   \caption{An illustration of our proposed color augmentation. The left hand side shows the generation of positive/negative samples by swapping pre-existing illuminants from a pair of images via estimated color mapping matrices $\mathbf{M}_{\mathrm{AB}}$ and $\mathbf{M}_{\mathrm{BA}}$. The right hand side shows the augmented samples with novel illuminants via interpolation ($w = +0.5$) and extrapolation ($w = -1.5$ and $w = +1.5$) using the detected color checkers $\mathbf{C}_{\mathrm{A}}$ and $\mathbf{C}_{\mathrm{B}}$.}
\label{fig:coloraug}
\end{figure*}
\subsection{Raw-domain Color Augmentation}
\label{sec:coloraug} 
The goal of our proposed color augmentation is to synthesize
more diverse and harder positive and negative samples by manipulating illuminants such that the color constancy solution space is better constrained.
%
As shown in Fig.~\ref{fig:coloraug}, for example, given two randomly sampled  $(\mathbf{I}_{\mathrm{XA}}, L_{\mathbf{A}})$, and $(\mathbf{I}_{\mathrm{YB}}, L_{\mathbf{B}})$ from training data, we go through the following procedure to synthesize  $\mathbf{I}_{\mathrm{YC}}$, $\mathbf{I}_{\mathrm{YA}}$ and $\mathbf{I}_{\mathrm{XC}}$, as defined in Section~\ref{sec:clcc}.
\paragraph{Color checker detection}
We extract 24 linear-raw RGB colors $\mathbf{C}_{\mathrm{A}} \in \mathbb{R}^{24 \times 3}$ and $\mathbf{C}_{\mathrm{B}} \in \mathbb{R}^{24 \times 3}$ of the color checker from $\mathbf{I}_{\mathrm{XA}}$ and $\mathbf{I}_{\mathrm{YB}}$ respectively using the off-the-shelf color checker detector.
\paragraph{Color transformation matrix}
Given $\mathbf{C}_{\mathrm{A}}$ and $\mathbf{C}_{\mathrm{B}}$, we can solve a linear mapping $\mathbf{M}_{\mathrm{AB}} \in \mathbb{R}^{3 \times 3}$ that transform $\mathbf{C}_{\mathrm{A}}$ to $\mathbf{C}_{\mathrm{B}}$ by any standard least-square method. The inverse mapping $\mathbf{M}_{\mathrm{BA}}$ can be derived by solving the $\mathbf{M}_{\mathrm{AB}}^{-1}$. Accordingly, we can augment $\mathbf{I}_{\mathrm{XB}}$ and $\mathbf{I}_{\mathrm{YA}}$ as:
\begin{equation}
\begin{aligned}
\mathbf{I}_{\mathrm{XB}} &= \mathbf{I}_{\mathrm{XA}} \mathbf{M}_{\mathrm{AB}}\\
\mathbf{I}_{\mathrm{YA}} &= \mathbf{I}_{\mathrm{YB}} \mathbf{M}_{\mathrm{BA}}
\end{aligned}
\end{equation}
\paragraph{Novel illuminant synthesis}
The above augmentation procedure produces novel samples $\mathbf{I}_{\mathrm{XB}}$ and $\mathbf{I}_{\mathrm{YA}}$, but using only pre-existing illuminants $L_{\mathrm{A}}$ and $L_{\mathrm{B}}$ from the training data. 
To synthesize a novel sample $\mathbf{I}_{\mathrm{XC}}$ under a novel illuminant $L_{\mathrm{C}}$ that \emph{does not exist in the training dataset}, 
we can synthesize $\mathbf{C}_{\mathrm{C}}$ by channel-wise \emph{interpolating} or \emph{extrapolating} from the existing $\mathbf{C}_{\mathrm{A}}$ and $\mathbf{C}_{\mathrm{B}}$ as:
\begin{equation}
\begin{aligned}
\mathbf{C}_{\mathrm{C}} = (1-w) \mathbf{C}_{\mathrm{A}} + w \mathbf{C}_{\mathrm{B}}
\end{aligned}
\label{eqn:C_C}
\end{equation}
where $w$ can be randomly sampled from a uniform distribution of an appropriate range $[w_{min}, w_{max}]$. Note that $w$ should not be close to zero in avoidance of yielding a false negative sample $\mathbf{I}_{\mathrm{XC}} = \mathbf{I}_{\mathrm{XA}}$ for contrastive learning.

To more realistically synthesize $\mathbf{I}_{\mathrm{XC}}$ (i.e., more accurate on chromatic colors), we need the full color transformation matrix $\mathbf{M}_{\mathrm{AC}}$ that maps $\mathbf{I}_{\mathrm{XA}}$ to $\mathbf{I}_{\mathrm{XC}}$:
\begin{equation}
\begin{aligned}
\mathbf{I}_\mathrm{XC} &= \mathbf{I}_{\mathrm{XA}} \mathbf{M}_{\mathrm{AC}}\\
\mathbf{I}_\mathrm{YC} &= \mathbf{I}_{\mathrm{YB}} \mathbf{M}_{\mathrm{BC}}
\end{aligned}
\label{eqn:I_AC}
\end{equation}
where $\mathbf{M}_{\mathrm{AC}}$ can be efficiently computed from the identity matrix $\mathbbm{1}$ and $\mathbf{M}_{\mathrm{AB}}$ without solving least-squares as:
\begin{equation}
\begin{aligned}
\mathbf{M}_\mathrm{AC} &= (1-w) \mathbbm{1} + w \mathbf{M}_{\mathrm{AB}}\\
\mathbf{M}_\mathrm{BC} &= w \mathbbm{1} + (1-w) \mathbf{M}_{\mathrm{BA}}\\
\end{aligned}
\label{eqn:M_AC}
\end{equation}
Equation \eqref{eqn:M_AC} can be derived from Eq.\eqref{eqn:C_C} and Eq.\eqref{eqn:I_AC}.
\paragraph{From full color mapping to neutral color mapping}
Our synthesis method could be limited by the performance of color checker detection.
When the color checker detection is not successful, the full colors $\mathbf{C}_{\mathrm{A}}$ and $\mathbf{C}_{\mathrm{B}}$ could be reduced to the neutral ones $L_{\mathrm{A}}$ and $L_{\mathrm{B}}$,
meaning that the color transformation matrix $\mathbf{M}_{\mathrm{AB}}$ is reduced from a full matrix to a diagonal matrix.
This is also equivalent to first perform WB on $\mathbf{I}_{\mathrm{A}}$ with $L_{\mathrm{A}}$, and subsequently perform an inverse WB with $L_{\mathrm{B}}$.

We provide the ablation study of this simplified version in our experiment, where we term the full color mapping as \textit{Full-Aug} and the simplified neutral color mapping as \textit{WB-Aug}.
We show that even though chromatic colors cannot be correctly mapped, \textit{WB-Aug} could still obtain performance improvement over the baseline.
\begin{figure}[htbp]
            \centering
            \subfloat[(a) On the NUS-8 dataset, CLCC achieves the best results and the most light-weighted model of all comparable methods.]{\includegraphics[width=0.45\textwidth]{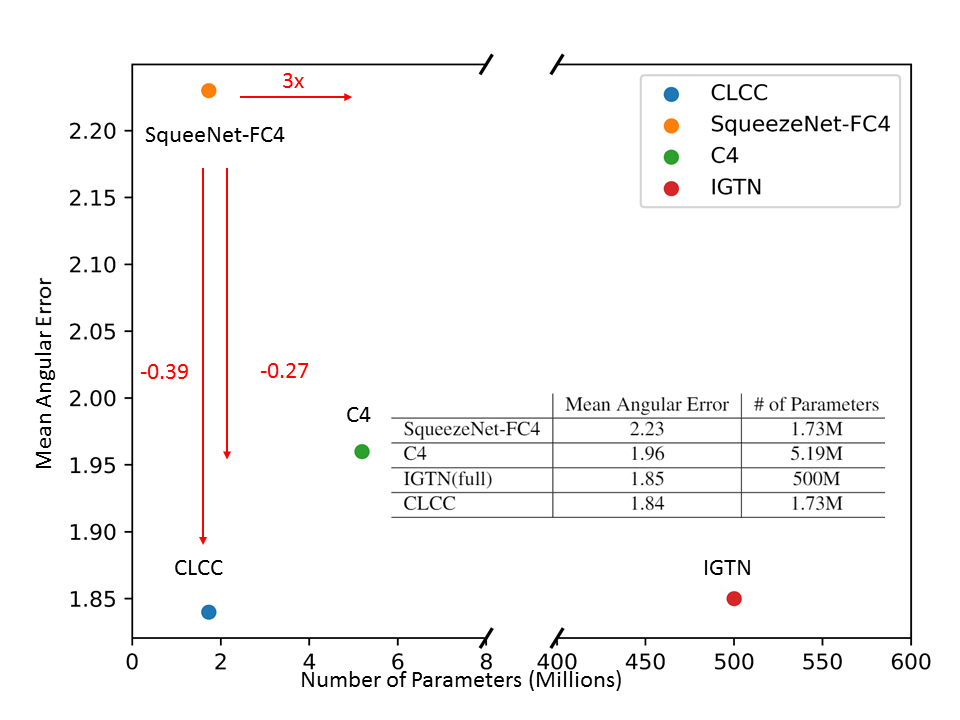}}
            \label{fig:NUS8Results}
            \\[-0.35em]
            \subfloat[(b) On the Gehler dataset, without increasing model complexity, CLCC improves SqueezeNet-FC4 to achieve comparable results.]{\includegraphics[width=0.45\textwidth]{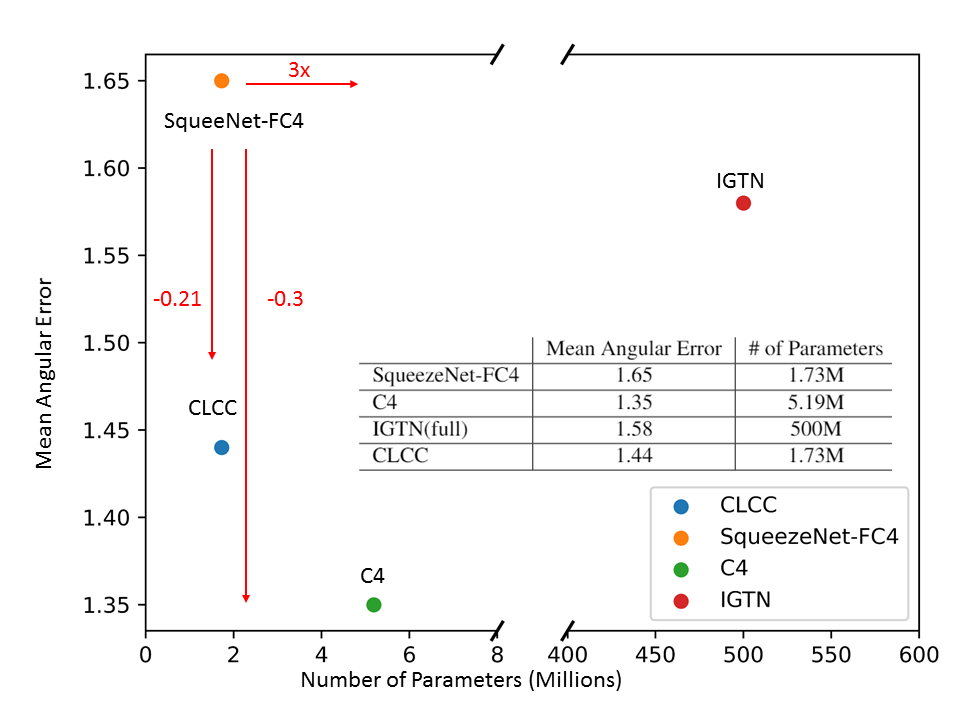}}
            \label{fig:GehlerResults}
            \\
            \caption{Model complexity versus mean angular error. CLCC improves SqueezeNet-FC4 by 0.39 (17.5\%) on the NUS-8 dataset as the state-of-the-art, and by 0.21 (12.5\%) on the Gehler dataset as a comparable method.}
            \label{fig:compare_with_other_methods}
\end{figure}

\section{Experiment}
\subsection{Network training}
Following FC4~\cite{HuWL17}, we use ImageNet-pretrained SqueezeNet as the backbone and add a non-linear projection head with three-layer MLP with $512$ hidden units for contrastive learning. Note that the projection head is thrown away once the learning is finished.
We use Adam~\cite{KingmaB14} optimizer with $\beta_1 = 0.9$ and $\beta_2 = 0.999$.
The learning rate is $0.0003$ and batch size is $16$.
We use dropout~\cite{SrivastavaHKSS14} with probability of $0.5$ and $L_2$ weight decay of $0.000057$ for regularization.
The loss weights for illuminant estimation and contrastive learning heads $(\lambda, \beta)$ is $(0.1, 1.0)$ for the first $5000$ epochs, $(1.0, 0.1)$ for the rest $5000$ epochs in learning objective~(\ref{eqn:learningobjective}).
The number of negative samples $N$ is $12$ and the temperature scaling factor $\tau$ is $0.87$ for InfoNCE loss~(\ref{eqn:infonceloss}).
Note that we do not train our illuminant estimation head with contrastive pairs. They are only used for training the contrastive learning head as depicted in Fig.~\ref{fig:overview_architecture}.
\subsection{Data augmentation}
We follow the default data augmentations used by FC4 with several differences. We resize the crop to $256\times256$ to speed-up training. For perturbation-based augmentations in contrastive learning, the range of intensity gain is $[0.8, 1.2]$, and the ranges of standard deviation of Guassian noise and shot noise are $[0, 0.04]$ and $[0.02, 0.06]$ for $[0,1]$-normalized images respectively. The $(w_{min}, w_{max})$ for novel color synthesis ~(\ref{eqn:C_C}) are $(-5.0, -0.3)$ and $(+0.3, +5.0)$.

\subsection{Dataset and evaluation metric}
There are two standard public datasets for color constancy task: the reprocessed~\cite{GehlerDataset} Color Checker Dataset~\cite{GehlerRBMS08} (termed as the Gehler dataset in this paper) and the NUS-8 Dataset~\cite{cheng2014illuminant}. The Gehler dataset has $568$ linear raw-RGB images captured by $2$ cameras and the NUS-8 dataset has $1736$ linear raw-RGB images captured by $8$ cameras.
In color constancy studies, three-fold cross validation is widely used for both datasets. Several standard metrics are reported in terms of angular error in degrees: mean, median, tri-mean of all the errors, mean of the lowest $25\%$ of errors
, and mean of the highest $25\%$ of errors.
%
\begin{center}
\begin{table}[t]
\resizebox{\columnwidth}{!}{
\begin{tabular}{ l c c c c c } 
\hline
           & Mean & Median & Tri. & Best-25\%      & Worst-25\%  \\
\hline
White-Patch~\cite{brainard1986analysis} & 10.62 & 10.58 & 10.49 & 1.86 & 19.45 \\
Edge-based Gamut~\cite{Barnard00} & 8.43 & 7.05 & 7.37 & 2.41 & 16.08 \\
Pixel-based Gamut~\cite{Barnard00} & 7.70 & 6.71 & 6.90 & 2.51 & 14.05 \\
Intersection-based Gamut~\cite{Barnard00} & 7.20 & 5.96 &  6.28 & 2.20 & 13.61 \\
Gray-World~\cite{buchsbaum1980spatial} & 4.14 & 3.20 &  3.39 & 0.90 & 9.00 \\
Bayesian~\cite{GehlerRBMS08} & 3.67 & 2.73 & 2.91 & 0.82 & 8.21 \\
NIS~\cite{GijsenijG11} & 3.71 & 2.60 & 2.84 & 0.79 & 8.47 \\
Shades-of-Gray~\cite{FinlaysonT04} & 3.40 & 2.57 & 2.73 & 0.77 & 7.41 \\
1st-order Gray-Edge~\cite{WeijerGG07} & 3.20 & 2.22 & 2.43 & 0.72 & 7.36 \\
2nd-order Gray-Edge~\cite{WeijerGG07} & 3.20 & 2.26 &  2.44 & 0.75 & 7.27 \\
Spatio-spectral (GenPrior)~\cite{ChakrabartiHZ12} & 2.96 & 2.33 & 2.47 & 0.80 & 6.18 \\
Corrected-Moment (Edge)~\cite{Finlayson13} & 3.03 & 2.11 & 2.25 & 0.68 & 7.08 \\
Corrected-Moment (Color)~\cite{Finlayson13} & 3.05 & 1.90 & 2.13 & 0.65 & 7.41 \\
Cheng et al.~\cite{cheng2014illuminant} & 2.92 & 2.04 & 2.24 & 0.62 & 6.61 \\
CCC (dist+ext)~\cite{Barron15} & 2.38 & 1.48 & 1.69 & 0.45 & 5.85 \\
Regression TreeTree~\cite{ChengPCB15} & 2.36 & 1.59 &  1.74 & 0.49 & 5.54 \\
DS-Net (HypNet + SelNet)~\cite{ShiLT16} & 2.24 & 1.46 & 1.68 & 0.48 & 6.08 \\
FFCC-4 channels~\cite{BarronT17} & 1.99 & \second{1.31} & \second{1.43} & \first{\textbf{0.35}} & 4.75 \\
\hline
AlexNet-FC4~\cite{HuWL17} & 2.12 & 1.53 & 1.67 & 0.48 & 4.78 \\
SqueezeNet-FC4~\cite{HuWL17} & 2.23 & 1.57 & 1.72 & 0.47 & 5.15 \\
IGTN (vanilla triplet loss)~\cite{XuLHLQ20} & 2.02 & 1.36 & - & 0.45 & 4.70 \\
IGTN (no triplet loss)~\cite{XuLHLQ20} & 2.28 & 1.64 & - & 0.51 & 5.20 \\
IGTN (no learnable histogram)~\cite{XuLHLQ20} & 2.15 & 1.52 & - & 0.47 & 5.28 \\
IGTN (full)~\cite{XuLHLQ20} & 1.85 & \first{\textbf{1.24}} & - & \second{0.36} & 4.58 \\
C4~\cite{YuCWQZJ20} & 1.96 & 1.42 &  1.53 &  0.48 & \second{4.40} \\
\hline
CLCC w/ Full-Aug & \first{\textbf{1.84}} & \second{1.31} & \first{\textbf{1.42}} & 0.41 & \first{\textbf{4.20}} \\
\hline
\end{tabular}
}
\caption{Angular error of various methods on the NUS-8 dataset. CLCC gets the best results on the mean tri-mean and worst-25\% metrics, and comparable results on the others. Notably, although IGTN gets the best result on the median metric, its model complexity is the largest.}
\label{NUS8Results}
\end{table}
\end{center}

\begin{center}
\begin{table}[t]
\resizebox{\columnwidth}{!}{
\begin{tabular}{ l c c c c c c } 
\hline
           & Mean & Median & Tri. & Best-25\%  & Worst-25\% & Extra data \\
\hline
Gray World~\cite{buchsbaum1980spatial} & 6.36 & 6.28 &  6.28 & 2.33    & 10.58  &   \\ 
General Gray World~\cite{WeijerGG07}   & 4.66 & 3.48 &  3.81 & 1.00    & 10.58  &   \\ 
White Patch~\cite{brainard1986analysis} & 7.55 & 5.68 & 6.35 & 1.45 & 16.12      &   \\
Shades-of-Gray~\cite{FinlaysonT04} & 4.93 & 4.01 &  4.23 & 1.14 & 10.20           &   \\
Spatio-spectral (GenPrior)~\cite{ChakrabartiHZ12} & 3.59 & 2.96 &  3.10 & 0.95 & 7.61 & \\
Cheng et al.~\cite{cheng2014illuminant} & 3.52 & 2.14 &  2.47 & 0.50 & 8.74 & \\
NIS~\cite{GijsenijG11} & 4.19 & 3.13 & 3.45 & 1.00 & 9.22 & \\
Corrected-Moment (Edge)~\cite{Finlayson13}& 3.12 & 2.38 & - & 0.90 & 6.46 & \\
Corrected-Moment (Color)~\cite{Finlayson13} & 2.96 & 2.15 & - & 0.64 & 6.69 & \\
Exemplar~\cite{JozeD12} & 3.10 & 2.30 & - & - & - & \\
Regression Tree~\cite{ChengPCB15} & 2.42 & 1.65 & 1.75 & 0.38 & 5.87 & \\
CNN~\cite{BiancoCS17} & 2.36 & 1.95 & - & - & - & \\
CCC (dist+ext)~\cite{Barron15} & 1.95 & 1.38 & 1.22 & 0.35 & 4.76 & \\
DS-Net(HypNet+SelNet)~\cite{ShiLT16} & 1.90 & 1.12 & 1.33 & 0.31 & 4.84 & \\
FFCC-4 channels~\cite{BarronT17} & 1.78 & 0.96 & 1.14 & 0.29 & 4.29 & \\
FFCC-2 channels~\cite{BarronT17} & 1.67 & 0.96 &  1.13 & 0.26 & 4.23 & +S \\
FFCC-2 channels~\cite{BarronT17} & 1.65 & \first{\textbf{0.86}} &  1.07 & 0.24 & 4.44 & +M\\
FFCC-2 channels~\cite{BarronT17} & 1.61 & \first{\textbf{0.86}} & \second{1.02} & \first{\textbf{0.23}} & 4.27 & +S +M\\
\hline
AlexNet-FC4~\cite{HuWL17} & 1.77 & 1.11 & 1.29 & 0.34 & 4.29 & \\
SqueezeNet-FC4~\cite{HuWL17} & 1.65 & 1.18 & 1.27 & 0.38 & 3.78 & \\
IGTN (vanilla triplet loss)~\cite{XuLHLQ20} & 1.73 & 1.09 & - & 0.31 & 4.25 & \\
IGTN (no triplet loss)~\cite{XuLHLQ20} & 1.78 & 1.13 & - & 0.34 & 4.31 & \\
IGTN (no learnable histogram)~\cite{XuLHLQ20} & 1.85 & 1.10 & - & 0.31 & 4.91 & \\
IGTN (full)~\cite{XuLHLQ20} & 1.58 & 0.92 & - & 0.28 & 3.70 & \\
C4 ~\cite{YuCWQZJ20} & \first{\textbf{1.35}} & \second{0.88} &  \first{\textbf{0.99}} & 0.28 & \first{\textbf{3.21}} & \\
\hline
CLCC w/ Full-Aug & \second{1.44} & 0.92 & 1.04 & \second{0.27} & \second{3.48} & \\
\hline
\end{tabular}
}
\caption{Angular error of various methods on the Gehler dataset. The use of semantic data or meta-data are denoted by “S” or "M". The result shows that SqueezeNet-FC4 plugging in our approach, which keeps the same model complexity and without meta data can achieve comparable performance.}
\label{GehlerResults}
\end{table}
\end{center}
\subsection{Evaluation}
%
\paragraph{Quantitative evaluation}
Following the evaluation protocol, we perform three-fold cross validation on the NUS-8 and the Gehler datasets.
We compare our performance with previous state-of-the-art approaches.
As shown in Fig.~\ref{fig:compare_with_other_methods}a, the proposed CLCC is able to achieve state-of-the-art mean angular error on the NUS-8 dataset, 17.5$\%$ improvements compared to FC4 with similar model size.  
Other competitive methods, such as C4 and IGTN, use much more model parameters ($3\times$ and more than $200\times$) but give worse mean angular error.
Table~\ref{NUS8Results} shows comprehensive performance comparisons with recent methods on the NUS-8 dataset~\cite{cheng2014illuminant}. Our CLCC provides significant improvements over the baseline network SqueezeNet-FC4 across all scoring metrics and reach the best mean metric, as well as the best worst-25\% metric. 
This indicates that the proposed fully-supervised contrastive learning not only improves the overall performance when there is no massive training data, but also improves robustness via effective constrastive pairs constructions. For the Gehler dataset, as shown in Fig.~\ref{fig:compare_with_other_methods}b, our CLCC stays competitive with less than $0.1$ performance gap behind the best performing approach C4~\cite{YuCWQZJ20}, whose model size is $3\times$ larger. Table~\ref{GehlerResults} shows detailed performance of state-of-the-art methods on the Gehler dataset. It is shown that methods achieving better scores than CLCC either require substantially more complexity (C4), or utilize supplemental data (FFCC).
C4 has three times more parameters which may facilitate remembering more sensor features than ours.
FFCC needs meta-data from camera to reach the best median metric. If no auxiliary data is used, CLCC performs better than FFCC-4 channels on all metrics.


\paragraph{Ablation for color augmentation}
Recap that our proposed color augmentation methods for contrastive learning includes Full-Aug and WB-Aug mentioned in Section~\ref{sec:coloraug}. As shown in Table~\ref{ablation_augmentation}, even when the color checker is not successfully detected for full color mapping (Full-Aug), the reduced neutral color mapping (WB-Aug) is still able to significantly decrease the mean angular error from $2.23$ to $1.93$ and the worst-case error from $5.15$ to $4.30$ over the SqueezeNet-FC4 baseline, which are substantial relative improvement $13.5\%$ and $16.5\%$ respectively. 
Furthermore, when Full-Aug is considered, the mean angular error can be decreased from $1.93$ to $1.84$ with an additional relative improvement $5.1\%$. This shows that correctly mapped chromatic colors for synthesizing contrastive pairs can improve the quality of contrastive learning, resulting a improved model. 
%
\begin{table}[th]
\centering
\resizebox{\columnwidth}{!}{
\begin{tabular}{lcccc} 
\hline
           & Mean & Median & Best-25\% & Worst-25\% \\
\hline
FC4~\cite{HuWL17} & 2.23 & 1.57 & 0.47 & 5.15 \\
$+$ CLCC w/ WB-Aug & 1.93 & 1.38 & 0.44 & 4.30 \\
$+$ CLCC w/ Full-Aug & 1.84 & 1.31 & 0.41 & 4.20 \\
\hline
\end{tabular}
}
\caption{The results show CLCC is able to improve SqueezeNet-FC4 quality by contrastive learning with two different data augmentations on the NUS-8 dataset.}
\label{ablation_augmentation}
\end{table}

\begin{figure}[th]
\begin{center}
    \includegraphics[width=\linewidth]{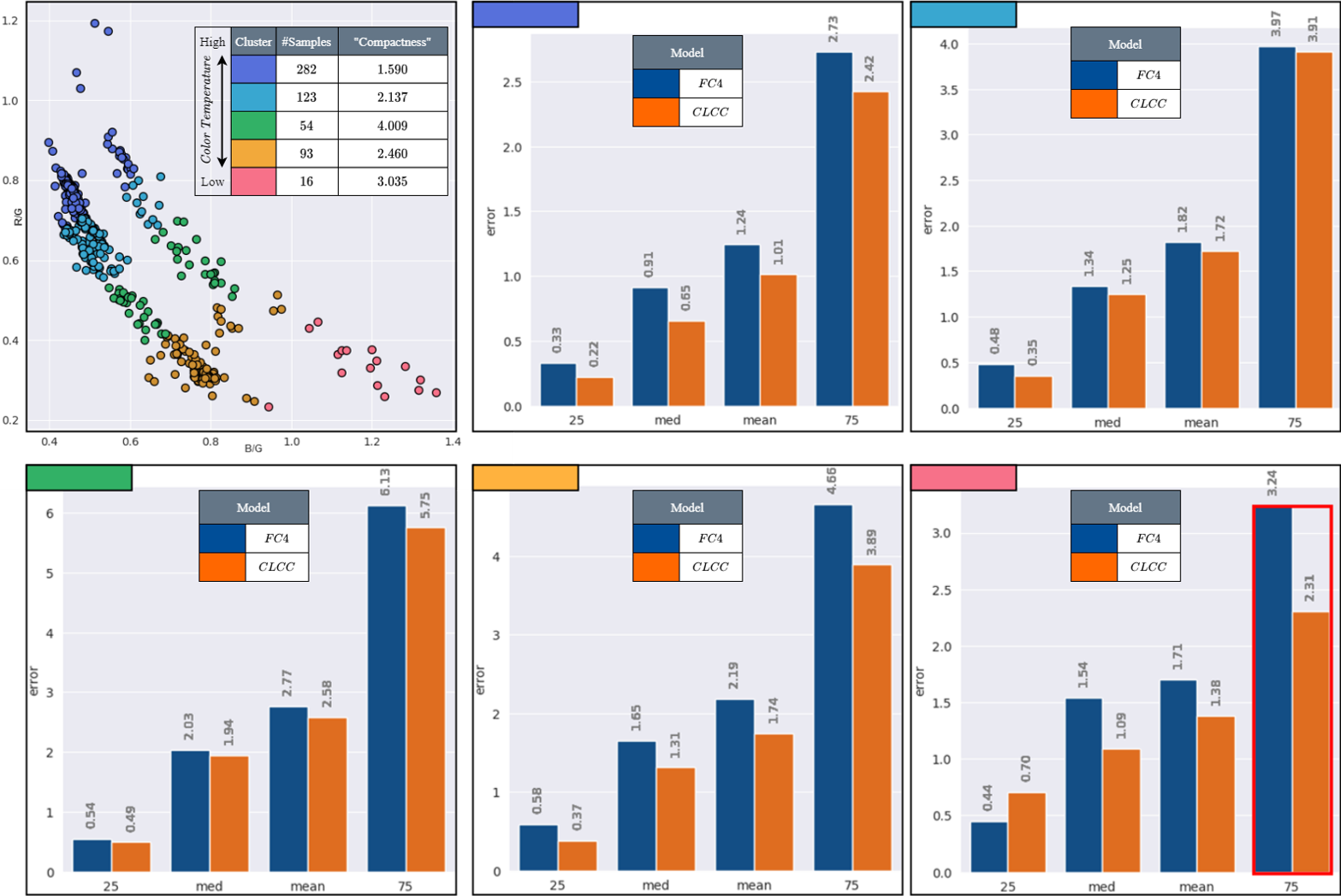}
\end{center}
   \caption{Per-cluster error metrics on the Gehler dataset. We show that CLCC achieves better performance on all clusters, especially the worst-case performance (i.e. worst-25\%). Notably, in the sparse data regime (cluster colored with \colorbox{pink!80}{pink} that contains only $16$ data points), CLCC trades best-case performance (i.e., best-25\%) with the worst-case one, leading to better robustness (i.e., lower test error standard deviation).}
\label{fig:quantitative_robustness}
\end{figure}
%


\paragraph{Worst-case robustness}
In this section, we are also interested in whether CLCC can provide improvements on robustness for worst-cases. To illustrate the robustness in more finegrained level, we propose to evaluate our model under $K$ grouped data on the Gehler dataset via clustering the illuminant labels with K-means. $K$ is selected as $5$ for example. Each group represents different scene contents under similar illuminants. As shown in Fig.~\ref{fig:quantitative_robustness}, CLCC greatly improves on all scoring metrics among all clusters (except for best-$25\%$ in pink cluster). Remarkably, we demonstrate that, when the amount of cluster data decreases from higher one (e.g., purple cluster) to lower one (e.g., pink cluster), as shown in the data distribution on top-left side in Fig.~\ref{fig:quantitative_robustness}), the improvement over worse-case performance increases. Especially in the region that suffers from data sparsity (e.g., $16$ data points in pink cluster), CLCC significantly reduces the worse-case error from $3.24$ to $2.31$, which achieves $28.7\%$ relative improvement.
This finding supports our contrastive learning design which aims to learn better illuminant-dependent features that are robust and invariant to scene contents.
\section{Conclusion}
In this paper, we present CLCC, a contrastive learning framework for color constancy.
Our framework differs from conventional self-supervised contrastive learning on the novel fully-supervised construction of contrastive pairs, driven by our novel color augmentation.
We improve considerably over previous strong baseline, achieving state-of-the-art or competitive results on two public benchmark datasets, without additional computational costs.
Our design of contrastive pairs allows model to learn better illuminant features that are particularly robust to worse-cases in data sparse regions.

{\small
\bibliographystyle{ieee_fullname}
\bibliography{egbib}
}

\end{document}